%% file: main.tex
\documentclass[10pt,twocolumn,letterpaper]{article}

\usepackage[pagenumbers]{cvpr} 
\usepackage{booktabs}
\usepackage{multirow}
\usepackage{algorithm}
\usepackage{algorithmic}
\usepackage{newfloat}
\usepackage{listings}
\usepackage{longtable}
\usepackage{amsthm,amsmath,amssymb}
\usepackage{mathrsfs}


\definecolor{cvprblue}{rgb}{0.21,0.49,0.74}

\usepackage[pagebackref,breaklinks,colorlinks,allcolors=cvprblue]{hyperref}

\def\paperID{17413} 
\def\confName{CVPR}
\def\confYear{2025}

\begin{document}

\title{Unlearning through Knowledge Overwriting: Reversible Federated Unlearning via Selective Sparse Adapter}

\author{
Zhengyi Zhong$^{1}$, \hspace{2pt} 
Weidong Bao$^{1}$, \hspace{2pt}
Ji Wang$^{1}$\thanks{Corresponding Author: Ji Wang}, \hspace{2pt}
Shuai Zhang$^{1}$, \hspace{2pt}
Jingxuan Zhou$^{1}$,
\\Lingjuan Lyu$^{2}$, \hspace{2pt}
Wei Yang Bryan Lim$^{3}$
\\ $^{1}$Laboratory for Big Data and Decision, National University of Defense Technology, China.
\\$^{2}$ Sony AI, Japan.$^{3}$ Nanyang Technological University, Singapore.
\\
$\rm{\{zhongzhengyi20, wdbao, wangji, zhangshuai20, zhoujingxuan\}@nudt.edu.cn}$, \\$\rm{lingjuanlvsmile@gmail.com, bryan.limwy@ntu.edu.sg}$\\
{\tt{\small{\url{https://github.com/Zhong-Zhengyi/FUSED-Code}}}}
}
\maketitle
\input{sec/0_abstract}    
\input{sec/1_intro}
\input{sec/2_relatedwork}
\input{sec/3_formulation}

\input{sec/4_method}

\input{sec/5_exp}
\input{sec/6_conclusion}

\noindent\textbf{Acknowledgement.}
{This work was supported in part by the National Natural Science Foundation of China under Grants 62102445 and 62002369, in part by the Postgraduate Scientific Research Innovation Project of Hunan Province under Grant CX20230075.}
{
    \small
    \bibliographystyle{ieeenat_fullname}
    \bibliography{main}
}


\end{document}

%% file: sec/0_abstract.tex
\begin{abstract}
Federated Learning is a promising paradigm for privacy-preserving collaborative model training. In practice, it is essential not only to continuously train the model to acquire new knowledge but also to guarantee old knowledge the \textit{right to be forgotten} (i.e., federated unlearning), especially for privacy-sensitive information or harmful knowledge. However, current federated unlearning methods face several challenges, including indiscriminate unlearning of cross-client knowledge, irreversibility of unlearning, and significant unlearning costs. To this end, we propose a method named \texttt{FUSED}, which first identifies critical layers by analyzing each layer’s sensitivity to knowledge and constructs sparse unlearning adapters for sensitive ones. Then, the adapters are trained without altering the original parameters, overwriting the unlearning knowledge with the remaining knowledge. This knowledge overwriting process enables \texttt{FUSED} to mitigate the effects of indiscriminate unlearning. Moreover, the introduction of independent adapters makes unlearning reversible and significantly reduces the unlearning costs. Finally, extensive experiments on three datasets across various unlearning scenarios demonstrate that \texttt{FUSED}'s effectiveness is comparable to Retraining, surpassing all other baselines while greatly reducing unlearning costs.
\end{abstract}

%% file: sec/1_intro.tex
\section{Introduction}
\label{sec:intro}

\textbf{Background.} Federated Learning (FL) \cite{McMahan2017} has emerged as a promising paradigm for privacy-preserving collaborative model training. In practice, FL models need to acquire new knowledge continuously while also ensuring the \textit{``right to be forgotten"} for previously used training data \cite{Villaronga2018, Garg2020}. For example, a year after the launch of ChatGPT, The New York Times accused OpenAI and Microsoft of the unauthorized use of its media data for training, demanding that they delete the acquired knowledge from its models \cite{CNN2023}. Furthermore, malicious clients may inject harmful data during training, potentially poisoning the global model. As a result, it is crucial for the global model to eliminate such harmful knowledge. This leads to the concept of \textit{Federated Unlearning} (FU).

\textbf{Challenges.} In the field of FU, two primary categories of methods have emerged: retraining-based methods \cite{Liu2022} and model manipulation-based methods \cite{Wang2022}. Among these, retraining-based methods are widely regarded as the state-of-the-art (SoTA) for achieving model unlearning. This approach involves removing the data designated for unlearning and retraining the model from scratch until convergence. Conversely, model manipulation methods modify the model directly using techniques such as gradient ascent, knowledge distillation, and setting parameters to zero to eliminate knowledge. However, existing methods still face several challenges:
\begin{itemize}
    \item \textit{Indiscriminate unlearning}: In scenarios where knowledge overlaps occur among clients, traditional methods indiscriminately remove shared knowledge during the unlearning process, leading to a substantial decline in the performance of other clients.
    \item \textit{Irreversible unlearning}: In FL systems, clients' unlearning requests may change dynamically. When a client no longer needs to forget certain knowledge, traditional methods cannot recover that memory quickly.
    \item \textit{Significant unlearning costs}: The retraining-based method requires multiple iterations, resulting in significant computational and communication costs. Even simple adjustments to model parameters can demand a significant amount of storage as a compensatory cost.
\end{itemize}

\textbf{Method.} To address these challenges, we propose a reversible \textbf{F}ederated \textbf{U}nlearning method via \textbf{SE}lective sparse a\textbf{D}apter (\texttt{FUSED}). To begin, we perform a layer-wise analysis of the model's sensitivity to knowledge changes, identifying the layers that are most affected. These sensitive layers are then processed into sparse structures known as unlearning adapters. This process, termed Critical Layer Identification (CLI), significantly reduces the number of model parameters, thereby lowering unlearning costs. Subsequently, the unlearning adapters are distributed to clients that do not require unlearning for retraining. During this phase, the original model is frozen, and only the independent unlearning adapters are trained. Ultimately, the unlearning adapters are integrated with the original model to yield a global unlearning model. This method leverages training on the remaining knowledge to effectively overwrite the knowledge that needs to be forgotten, addressing the issue of indiscriminate unlearning. Moreover, the introduction of independent adapters facilitates rapid recovery of forgotten knowledge through their removal and significantly reduces unlearning costs by utilizing sparse parameters. In summary, \texttt{FUSED} achieves high performance, reversibility, and cost-efficiency in FU, making it suitable for scenarios involving client unlearning, class unlearning, and sample unlearning scenarios.

\textbf{Contributions.} The contributions are as follows:
\begin{itemize}
    \item We propose \texttt{FUSED}, a reversible FU approach that retrains independent sparse adapters for unlearning. These adapters effectively mitigate unlearning interference while ensuring that the unlearning is reversible.
    \item We introduce the CLI method which accurately identifies the model layers sensitive to knowledge changes and constructs sparse unlearning adapters, resulting in a significant reduction in parameter scale and lowering unlearning costs.
    \item We theoretically and experimentally prove the effectiveness of the proposed method across different unlearning scenarios in FL, including client unlearning, class unlearning, and sample unlearning.
\end{itemize}

%% file: sec/2_relatedwork.tex
\section{Related work}
\label{sec:formatting}

\textbf{Machine unlearning.} Currently, most researchers focus on machine unlearning (MU) within centralized scenarios \cite{Liu2024, Chen2024, Zhang2023}. Mainstream methods can be classified into two categories: data manipulation and model manipulation \cite{Nguyen2022}. Data manipulation includes data mixing and data partitioning. The former fine-tunes the model to forget specific samples by introducing interference data or by replacing existing data \cite{ Sinha2020, Tarun2023, Zhang2022, Felps2020, Graves2021}. In contrast, the latter divides the training dataset into multiple subsets and retrains only the subset that contains the data to be forgotten \cite{Bourtoule2021, Chen2022, Chen2022a, He2021, Ren2021}. Model manipulation \cite{Li2024, Chowdhury2024} contains three strategies: model transformation, model pruning, and model replacement. Model transformation methods directly update the model parameters to offset the influence of forgotten samples on the model \cite{Golatkar2020, Guo2019, Izzo2021, Warnecke2021}. Model pruning methods involve pruning from the original model \cite{Liu2021, Wang2022, Baumhauer2022, Guo2020}. Model replacement methods compute nearly all possible sub-models and store them alongside the deployed model. When an unlearning request is received, only the sub-models affected by the unlearning operation need to be replaced. This method is commonly utilized in machine learning models such as decision trees \cite{Schelter2021, Brophy2021, Wu2020}.

\textbf{Federated unlearning.} Unlike centralized unlearning, FU \cite{Liu2020} expands the unlearning objectives to client unlearning \cite{Su2023}, sample unlearning, and class unlearning \cite{Wu2022, Gu2024}. In this context, commonly used unlearning methods can be classified into retraining-based methods \cite{Liu2022} and parameter manipulation-based methods \cite{Cao2023, Wang2023, Halimi2022, Che2023}. Retraining-based methods means training a new model from scratch without unlearning data. For example, when a particular client needs to be forgotten, \cite{Liu2021} have proposed approaches that retrain the remaining clients to obtain corrected gradient directions, which are then used to update the global model stored on the server. \cite{Liu2022} utilized an improved quasi-Newton method to accelerate the training process. \cite{Su2023} reduces the time and computational resources required for retraining through clustering. Despite these efforts to mitigate the resource costs associated with retraining, the expenses remain unacceptable in real-world scenarios. Consequently, some researchers have proposed parameter manipulation-based unlearning methods. For instance, \cite{Wang2022} focuses on classification tasks using CNN models and achieves unlearning classes by pruning class-related channel parameters. Furthermore, \cite{Wu2022a} eliminates the contribution of a target client by subtracting the accumulated historical updates from the global model. It then uses the old global model as a teacher model to train the unlearning model, employing knowledge distillation techniques to restore the model's performance.

Overall, current research on FU is still limited, primarily focusing on client unlearning. Additionally, the issues of knowledge interference and irreversibility have not been adequately considered.

%% file: sec/3_formulation.tex
\section{Problem formulation}

\textbf{Centralized machine unlearning.}\label{Problem Formulation:1} We denote $\mathcal{D}^{u}$ as the data to be forgotten, and $\mathcal{D}$ as the entire training dataset, $\mathcal{D}=({x_i},{y_i} )_{i = 1}^n$. Then, $\mathcal{D}^r = {\mathcal{D}\backslash {\mathcal{D}^u}}$ represents the data to be retained. Let $\mathcal{M}^{r}$ denote the model before unlearning,  $\mathcal{M}^f$ is the model after unlearning, and $\mathcal{FGT}$(·) denote the unlearning process. The unlearning can be represented as:
\begin{equation}
    {\mathcal{M}^f} = {\mathcal{FGT}}({\mathcal{M}^r},\mathcal{D}^r,{\mathcal{D}^{u}}).
\end{equation}

The objectives of FU are threefold: (a) minimizing the performance of $\mathcal{M}^f$ on $\mathcal{D}^{u}$; (b) maximizing the performance on $\mathcal{D}^{r}$, and (c) minimizing the resources consumed by the unlearning process. Denoting $\mathcal{F (\cdot)}$ as the model test loss and $\mathcal{RC (\cdot)}$ as resource consumption, the above objectives can be respectively expressed as:
\begin{equation}
\max {\mathcal{F}}({{\mathcal{M}}^f},({x_i},{y_i})),( {{x_i},{y_i}} ) \in {\mathcal{D}^u},
\end{equation}
\begin{equation}
\min {\mathcal{F}}({\mathcal{M}^f},({x_i},{y_i})),( {{x_i},{y_i}} ) \in {\mathcal{D}^r =  \mathcal{D}\backslash {\mathcal{D}^u}},
\end{equation}
\begin{equation}
\min {\mathcal{RC}}({\mathcal{FGT}}(\mathcal{M}^r, \mathcal{D}^r, \mathcal{D}^u)).
\end{equation}

Ideally, when a model is considered to have fully forgotten target knowledge, its performance should be equivalent to that of a model trained from scratch without ever seeing the forgotten data $\mathcal{D}^{u}$. In this retraining approach, it ensures the worst performance on the forgotten data $\mathcal{D}^{u}$ and the best performance on the remaining data $\mathcal{D}^{r}$. However, this approach requires significant computational resources and preserving all historical training data, which is impractical in real-world scenarios. Therefore, we posit that the closer the performance of the model $\mathcal{M}^f$ on $\mathcal{D}^{r}$ and $\mathcal{D}^{u}$ is to that of a retrained model, the better the unlearning effect, while also striving to minimize resource expenditure on this basis.

\textbf{Unlearning scenarios in FL.}\label{Problem Formulation:2} In consideration of the distributed nature of FL, traditional machine unlearning can be extended to client unlearning, class unlearning, and sample unlearning. In the case of client unlearning, we consider $N$ clients, a set of unlearning clients $N_u$, with the unlearning dataset $\mathcal{D}^{u}= \{\mathcal{D}_k\}_{k\in{N_u}}$, and remember dataset $\mathcal{D}^{r}= \{\mathcal{D}_k\}_{k\in{N\backslash{N_u}}}$, where $\mathcal{D}_k$ represents the data of client $k$. The optimization objectives are:
\begin{equation}
\max \sum\limits_{k \in {N_u}} {{\mathcal{F}}({\mathcal{M}^f},{\mathcal{D}_k})}, 
\end{equation}
\begin{equation}
\min \sum\limits_{k \in N\backslash {N_u}} {{\mathcal{F}}({\mathcal{M}^r},{\mathcal{D}_k})},
\end{equation}
\begin{equation}
\min {\mathcal{RC}}({\mathcal{FGT}}(\mathcal{M}^r, \{\mathcal{D}_k \}_{k \in N}).
\end{equation}
 
Sample unlearning means forgetting a portion of data within a client. It is similar to client unlearning. In the context of class unlearning, let all client data classes be $\mathcal{C}$ and the classes to be unlearned be $\mathcal{C}^u$. The unlearning dataset can be represented as ${\mathcal{D}^u} = \{ ( {x_i^k,y_i^k = c} ) \}_{c \in {\mathcal{C}^u},( {x_i^k,y_i^k} ) \in {\mathcal{D}_k},k \in N}$, and the remember dataset as ${\mathcal{D}^r} = {\{ {{\mathcal{D}_k}} \}_{k \in N}}\backslash {\mathcal{D}^u}$. The optimization objectives are:

\begin{equation}
\max \sum\limits_{({x_i^{.},y_i^{.}}) \in \{ {\mathcal{D}_k} \}_{k \in N}} {\mathcal{F}({\mathcal{M}^f},{{\left. {(x_i^{.},y_i^{.})} \right|}_{y_i^{.} \in {\mathcal{C}^u}}})},
\end{equation}
\begin{equation}
\min \sum\limits_{({x_i^{.},y_i^{.}}) \in \{ {\mathcal{D}_k}\}_{k \in N} } {\mathcal{F}({\mathcal{M}^r},{{\left. {(x_i^{.},y_i^{.})} \right|}_{y_i^{.} \notin {\mathcal{C}^u}}})},
\end{equation}
\begin{equation}
\min {\mathcal{RC}}({\mathcal{FGT}}(\mathcal{M}^r, {{\left. {(x_i^{.},y_i^{.})} \right|}_{y_i^{.}\in \mathcal{C}}})).
\end{equation}

\begin{figure*}[!ht]
	\centering
	\includegraphics[width=6.5in]{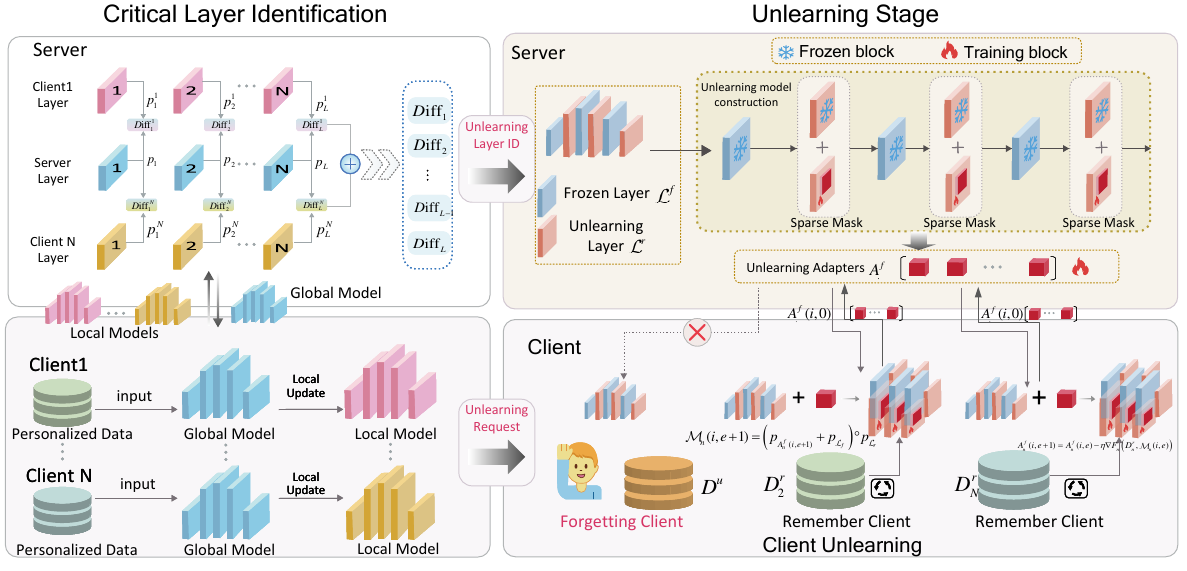}
	\caption{The figure illustrates the process of CLI (left) and unlearning (right). Left: the server computes the difference of each layer between the models uploaded by each client and the distributed one, identifying critical layers that are sensitive to knowledge. Right: a sparse adapter is constructed for each key layer, which is then independently trained on the remaining data.}
	\label{method}
\end{figure*}

%% file: sec/4_method.tex
\section{The proposed method: \texttt{FUSED}}
$\texttt{FUSED}$ involves a two-stage unlearning process (as shown in \cref{method}). The first stage is Critical Layer Identification (CLI), and the second stage is Unlearning via Sparse Adapters, which is based on the critical layers identified. 

\subsection{Critical layer identification}
During the CLI phase, each client, coordinated by the server, participates in a federated iteration process. Clients receive the global model distributed by the server and train it using their local data before uploading it back to the server. Subsequently, the distance between the parameters of each layer in the models from different clients and those in the corresponding layers of the initial model is calculated by the server. The layers with the most significant parameter changes are obtained by averaging these distances.

Consider a global model with $L$ layers, and $N$ clients, each with an identical model structure. After local training, the parameters of these models differ across clients. Let $p_l^n$ represent the parameters of the $l$-th layer of the $n$-th client, where $n = 1,2, \cdots ,N$ and $l = 1,2, \cdots ,L$. The initial distributed global model is denoted as $\mathcal{M}^r = \left\{ {{p_1},{p_2}, \cdots ,{p_L}} \right\}$. After local training by the clients, the variation in the $l$-th layer of the model can be expressed as:
\begin{equation}
    Diff_l = Diff_l^1(p_l^1,{p_l}) \oplus  \cdots  \oplus Diff_l^N(p_l^N,{p_l}),
    \label{eq1}
\end{equation}
where $Diff_l^n(p_l^n,{p_l})$ represents the difference between the $l$-th layer of the $n$-th client’s model and the $l$-th layer of the original model (need to be forgotten) distributed by the server. We utilize the Manhattan distance for measurement. Assuming that the dimensions of $p_l^n$ and $p_l$ are $k \times v$. The calculation process is as follows:
\begin{equation}
{\cal D}iff(p_l^n,{p_l}) = \sum\nolimits_{i = 1}^k {\sum\nolimits_{j = 1}^v {\left| {p_{l,ij}^n - {p_{l,ij}}} \right|} }.
    \label{eq2}
\end{equation}

The aggregation method of $\oplus$ is as follows:
\begin{equation}
\begin{array}{l}
Diff_l^1(p_l^1,{p_l}) \oplus {\cal D}iff_l^N(p_l^N,{p_l})= \frac{{\left| {{D_1}} \right|}}{{\sum\nolimits_{n = 1}^N {\left| {{D_n}} \right|} }}\\
 Diff_l^1(p_l^1,{p_l}) + \frac{{\left| {{D_1}} \right|}}{{\sum\nolimits_{n = 1}^N {\left| {{D_n}} \right|} }}{\cal D}iff_l^N(p_l^N,{p_l}),
\end{array}
\label{eq3}
\end{equation}
where $\left|D_i\right|$ represents the data volume of client $i$. Eventually, $LS = \left[ {\mathop {\arg \max }\limits_l \{ Diff_l\} , \cdots ,\mathop {\arg \min }\limits_l \{ Diff_l\} } \right]$ indicating the changes in different model layers is obtained. The first element corresponds to the layer index that is most sensitive to changes in client knowledge, while the last element corresponds to the most robust layer. To minimize resource cost, the subsequent unlearning process prioritizes unlearning in the most sensitive model layers.

\subsection{Unlearning via sparse adapters}
Based on the list obtained from CLI, given a preset value $K$ for the number of layers to be unlearned, the first $K$ indices in $LS$ are designated for FU. Let $\mathcal{L}^f$ denote the set of layers that need to be unlearned, and $\mathcal{L}^r$ denote the remaining frozen layers. For each unlearning layer in $\mathcal{L}^f$, we discard most of the parameters in a random manner and leave only a small portion, forming a sparse parameter matrix ${A^f}$. During training, only ${A^f}$ is trained while $\mathcal{L}^f$ remains unchanged. In the inference process, the new parameters of the unlearning layer are directly obtained by adding the parameters of ${A^f}$ (denoted by ${p_{{A^f}}}$) and $\mathcal{L}^f$ (denoted by ${p_{{{\mathcal{L}}^f}}}$).

For the original model $\mathcal{M}^r$, the entire unlearning process can be divided into four stages: model distribution, local training, model uploading, and model aggregation. In \texttt{FUSED}, there are significant differences in the model distribution and local training stages compared to traditional FL. The following sections will primarily focus on these two stages. Firstly, during the model distribution stage, the model is only distributed to clients that contain the remembered dataset $\mathcal{D}^r$ (see Problem Formulation), and only unlearning adapters are transmitted. This means that in client unlearning scenarios, the clients to be forgotten will not receive the adapters\footnote{In class or sample unlearning, the classes/samples that need to be forgotten will no longer participate in training with the server.}. Additionally, the distributed model is a sparse matrix ${A^f}$, which significantly reduces communication overhead. Secondly, in the local training stage, for a client $n$, assuming the total number of local training epochs per federated iteration is $E$, then in the $t$-th federated iteration, the parameters of the model $\mathcal{M}_n(i,e)$ are $({p_{A_n^f(i,e)}} + {p_{{{\mathcal{L}}^f}}}) \circ {p_{{{\mathcal{L}}^r}}}$. The training process is as follows:
\begin{equation}
    A_n^f(i,e + 1) = A_n^f(i,e) - \eta \nabla {F_n}(D_n^r,{{\cal M}_n}(i,e)),
    \label{eq4}
\end{equation}

\begin{equation}
    {{\mathcal{M}}_n}(i,e + 1) = ({p_{A_n^f(i,e + 1)}} + {p_{{{\cal L}_f}}}) \circ {p_{{{\cal L}_r}}},
    \label{eq5}
\end{equation}
where $e = 0, \cdots E - 1$, ${F_n}(D_n^r,{\mathcal{M}_n}(i,e))$ represents the loss and $\eta$ denotes the learning rate. In each round of local training, $\mathcal{M}_n(i,e)$ is derived from the fusion of the original model $\mathcal{M}^r$ and the sparse matrix $A_n^f(i,e)$ obtained from the previous round. Each completed training round corresponds to a process of knowledge overwriting, during which the remaining knowledge is progressively enhanced. It is worth noting that during the training process, only ${p_{A_n^f(i,e)}}$ is updated. The other parameters, ${p_{{{\mathcal{L}}^f}}}$ and ${p_{{{\mathcal{L}}^r}}}$, remain frozen and are only used to compute the loss during inference. After local training is completed, each client uploads ${p_{A_n^f(I, E)}}$ to the server, which aggregates the updates using the FedAvg \cite{McMahan2017} method to obtain a new ${p_{A_n^f(i+1)}}$. After training, we need to concatenate the adapter with the corresponding unlearning layer of the original model to derive the global unlearning model. When the client's knowledge no longer needs to be unlearned, removing the unlearning adapters will effectively restore the original memory, thereby making the unlearning process reversible.

\subsection{Algorithm}
To elucidate the aforementioned training process more clearly, this section presents it through pseudocode (as shown in \cref{Federated Unlearning}). The inputs for \cref{Federated Unlearning} are the original model to be unlearned $\mathcal{M}^r$, the total number of federated training rounds $I$, the number of local training epochs $E$, the number of clients $N$, the forgetting dataset $\mathcal{D}^u$, and the dataset for all clients $\mathcal{D}$. The output is the model $\mathcal{M}^f$ after the unlearning process. First, based on the original model $\mathcal{M}^r$, a federated iteration is conducted to determine the model layers to be unlearned (details of CLI are displayed in the Supplementary Material). Using the indexes of unlearning layers, a series of unlearning adapters are constructed through random sparsification. Then, the FU process begins: during the first federated iteration, the server distributes both the unlearning adapters and the original model to the clients that contain the remembered dataset (lines 4-7 in \cref{Federated Unlearning}), excluding clients that only contain the forgetting dataset. Subsequently, the clients freeze the original model (line 9 in \cref{Federated Unlearning}) and merge the unlearning adapter with the original model (line 11 in \cref{Federated Unlearning}). The clients then train unlearning adapters using local data and upload it to the server for aggregation (line 16 in Algorithm 1). After $I$ rounds of federated iterations, the final unlearning adapters are merged with the original model (line 18 in \cref{Federated Unlearning}) to obtain the global unlearned model $\mathcal{M}^f$.

\begin{algorithm}[tb]
\caption{Our method \texttt{FUSED}}
\label{Federated Unlearning}
\textbf{Input}: Original model $\mathcal{M}^r$, number of iteration $I$, number of local training $E$, number of clients $N$, unlearning data $\mathcal{D}_u$, all clients' data $\mathcal{D}$\\
\textbf{Output}: Unlearning model $\mathcal{M}^f$
\begin{algorithmic}[1] 
\STATE $\mathcal{L}^f$ $\gets$ $Unlearning$ $Layer$ $Identification$
\STATE Adapters $A^f$ $\gets$ Random dropout parameters of $\mathcal{L}_f$
\FOR{iteration $i=0$ \TO $I$}
\IF{i=0}
\STATE Server distribute $\mathcal{M}^r$ to all clients\\
\ENDIF
\STATE Server distribute adapters $A_f$ to clients maintaining $\mathcal{D}_r = {\mathcal{D}\backslash {\mathcal{D}_u}}$\\
\FOR{client $n = 1$ \TO $N$}
\STATE Freezing the original parameter $\mathcal{M}^r$
\FOR{ local epoch $e = 0$ \TO $E-1$}
\STATE Merge the ${\mathcal{M}^r}$ model with adapters to get ${\mathcal{M}_n}(e)$ (refer to \cref{eq5})
\STATE Update adapters to get $A^f_n(e+1)$ (\cref{eq4})
\ENDFOR
\STATE Upload $A^f_n(E)$ to the server
\ENDFOR
\STATE Server aggregates adapters to get $A^f$
\ENDFOR
\STATE Merge ${\mathcal{M}^r}$ with $A^f$ to get $\mathcal{M}^f$ 
\STATE \textbf{return} Unlearning model $\mathcal{M}^f$
\end{algorithmic}
\end{algorithm}

\subsection{Theoretical analysis}

In this section, we will theoretically analyze how knowledge overwriting happens in the training process of different tasks, providing theoretical support for the \texttt{FUSED}.

Suppose there are two learning tasks denoted as $T_1$ and $T_2$, each task has an input space $X$ and an output space $Y$. The parameterized model is represented as a mapping $f(\Theta):X\rightarrow Y$, $\Theta$ denotes an $n$-dimensional parameter vector of the neural network. For each task, we use a loss function $\mathcal{L}_{T}(\Theta)$ to measure the performance of the model.

The model learns on tasks $T_1$ and $T_2$ and obtains the optimized parameters $\Theta_1^*$ and $\Theta_2^*$, which are, respectively, $\epsilon_1$ and $\epsilon_2$ away from the minimum value, where $\epsilon_1$ and $\epsilon_2$ are arbitrarily small positive numbers:
\begin{equation}
\Theta_1^*=\mathop{\arg \min}\limits_{\Theta}(\mathcal{L}_{T_1}(\Theta)+\epsilon_1),~\epsilon_1>0.
    \label{eq6}
\end{equation}
\begin{equation}
\Theta_2^*=\mathop{\arg\min}\limits_{\Theta}(\mathcal{L}_{T_2}(\Theta)+\epsilon_2),~\epsilon_2>0.
    \label{eq7}
\end{equation}

We have the following assumptions:

\textit{\textbf{Assumption 1: }The loss function $\mathcal{L}_T(\Theta)$ is continuous and differentiable with respect to the parameters $\Theta$.}

\textit{\textbf{Assumption 2:} Near the optimized parameters $\Theta_1^*$, the loss function $\mathcal{L}_{T_1}(\Theta)$ can be approximated using a first-order Taylor expansion.}

From the work of \cite{Lee2019}, we notice the agreement between the predictions of the original network and those of the linear model obtained from the first-order Taylor expansion of the network.
Consider the linear model of the loss function $\mathcal{L}_{T_1}(\Theta_2^*)$ for the old task $T_1$ at the optimized parameters $\Theta_1^*$:
\begin{equation}
\mathcal{L}_{T_1}(\Theta_2^*)\approx \mathcal{L}_{T_1}(\Theta_1^*)+\nabla_\Theta \mathcal{L}_{T_1}(\Theta_1^*)^{\mathrm{T}}\cdot(\Theta_2^*-\Theta_1^*).
    \label{eq9}
\end{equation}

Performance degradation $\Delta \mathcal{L}_{T_1}$ is defined as:
\begin{equation}
\Delta \mathcal{L}_{T_1}\!=\!\mathcal{L}_{T_1}\!(\Theta_2^*)\!-\!\mathcal{L}_{T_1}\!(\Theta_1^*)\!\approx\!\nabla_\Theta\!\mathcal{L}_{T_1}\!(\Theta_1^*)\!^{\mathrm{T}}(\Theta_2^*\!-\!\Theta_1^*).
\label{eq10}
\end{equation}

Defining the task difference $\Phi(\Theta, T_1, T_2)$ as the cosine similarity between the gradients of two tasks.At the optimized parameters $\Theta_1^*$, we have:
\begin{equation}
\Phi(\Theta_1^*,T_1,T_2)=\frac{\nabla_\Theta \mathcal{L}_{T_1}(\Theta_1^*)^{\mathrm{T}}\cdot\nabla_\Theta \mathcal{L}_{T_2}(\Theta_1^*)}{||\nabla_\Theta \mathcal{L}_{T_1}(\Theta_1^*)||\cdot||\nabla_\Theta \mathcal{L}_{T_2}(\Theta_1^*)||}.
    \label{eq13}
\end{equation}

The changing direction of the parameters $\Theta$ is influenced by the gradient of the new task $T_2$, that is:
\begin{equation}
\Theta_2^*-\Theta_1^*=-\eta \nabla_\Theta \mathcal{L}_{T_2}(\Theta_1^*),
    \label{eq14}
\end{equation}
where $\eta$ represents the learning rate.

Substituting into the performance degradation approximation formula, we can obtain:
\begin{equation}
\Delta\mathcal{L}_{T_1}\!\approx\!-\eta\|\!\nabla_{\!\Theta}\!\mathcal{L}_{T_1}\!(\Theta_1^*)\!\|\|\!\nabla_{\!\Theta}\!\mathcal{L}_{T_2}\!(\Theta_1^*)\!\|\Phi(\Theta_1^*\!,T_1\!,T_2).
\label{eq15}
\end{equation}

When the cosine similarity of the gradients of the loss function at tasks $T_1$ and $T_2$ is less than $0$, the performance degradation of the old task is greater than $0$. That is, when the knowledge of the tasks learned sequentially is opposite, the learning of new knowledge will overwrite the old knowledge that the model has mastered.

The aforementioned analysis is based on full training of model parameters. Specifically, in \texttt{FUSED}, only partial parameters of critical layers are updated with new data. Then, the result will be:
\begin{equation}
\begin{array}{l}
\Theta_2^*-\Theta_1^*=-\eta\cdot (\mathbf{v}\circ\nabla_\Theta \mathcal{L}_{T_2}(\Theta_1^*)),\\
\{\mathbf{v}\in\{0,1\}^n|P(v_i=1)=\mathrm{p},i\in\{1,\dots,n\}\},
\end{array}
    \label{eq16}
\end{equation}
$\mathbf{v}$ is an $n$-dimensional binary vector. Each element of $\mathbf{v}$ takes the value of $1$ with probability $p$. Then, we can get:
\begin{equation}
\begin{aligned}
    \mathbb{E}(\Delta \mathcal{L}_{T_1})=&
-\eta\cdot \mathrm{p}\cdot||\nabla_\Theta \mathcal{L}_{T_1}(\Theta_1^*)||\cdot\\
& ||\nabla_\Theta \mathcal{L}_{T_2}(\Theta_1^*)||\cdot\Phi(\Theta_1^*,T_1,T_2).
\end{aligned}
\label{eq17}
\end{equation}

Therefore, it can be inferred from the above formula that when the angle between the gradients of the new task and the old task is greater than 90 degrees, the old knowledge learned by the model can also be covered by training some parameters on the new task.


%% file: sec/5_exp.tex
\section{Experiments}
\subsection{Experimental setting}
The experiments are built on PyTorch 2.2.0, developing an FL framework comprising one server and 50 clients. The hardware environment uses an NVIDIA RTX 4090. Optimizers consisting of SGD and Adam are employed with a batch size of 128. The result is listed in \cref{results}.

The datasets include FashionMNIST \cite{Xiao2017}, Cifar10 and Cifar100 \cite{Krizhevsky2009}. We use the Dirichlet function to partition the dataset and conduct tests under two conditions: $\alpha = 1.0$ and $\alpha = 0.1$. In \cref{results}, we primarily present the results for $\alpha = 1.0$; for the results under non-independent and identically distributed, please refer to the appendix. The model used for FashionMNIST is LeNet, while ResNet18 is employed for training on Cifar100. For Cifar10, training is conducted using both ResNet18 and a vision-based Transformer model called SimpleViT \cite{Yoshioka2024}. Baselines include Retraining, Federaser \cite{Liu2021}, Exact-Fun \cite{Xiong2023}, and EraseClient \cite{Halimi2022}. Among these, Retraining is the upper bound of unlearning; it can achieve the effect of the model having never encountered the forgotten data.

\textbf{Evaluations.} 
We evaluate \texttt{FUSED} from multiple perspectives: 

\begin{itemize}
    \item \textbf{\textit{RA \& FA}}: \textit{RA} is the testing accuracy on the remaining dataset, which should be as high as possible to minimize knowledge interference. \textit{FA} is the testing accuracy on the unlearned dataset, which should be as low as possible.

    \item \textbf{\textit{Comp \& Comm}}: \textit{Comp} is the time to complete pre-defined unlearning iterations; \textit{Comm} denotes the data volume transmitted between a single client and the server.

    \item \textbf{\textit{MIA}}: the privacy leakage rate after unlearning, which is assessed in the context of membership inference attacks. Assuming that the forgotten data is private, a lower inference accuracy of the attack model indicates less privacy leakage. 

    \item \textbf{\textit{ReA}}: the accuracy of relearning, which refers to the accuracy that can be achieved after a specified number of iterations to relearn the unlearned knowledge. If the unlearned knowledge is effectively erased, the accuracy achieved during the subsequent relearning will be lower.
\end{itemize}

%

\begin{table*}[htbp]
  \centering
  \resizebox{\textwidth}{!}{
    \begin{tabular}{c|ccccc|cc|ccc}
    \toprule
          & \multicolumn{5}{c|}{\textit{Client Unlearning}} & \multicolumn{2}{c|}{\textit{Class Unlearning}} & \multicolumn{3}{c}{\textit{Sample Unlearning}} \\
\cmidrule{2-11}          & Retrain & Federaser & E-F & \texttt{FUSED} & E-C & Retrain & \texttt{FUSED} &      & Retrain & \texttt{FUSED} \\
\cmidrule{2-11}          & \multicolumn{10}{c}{\textit{FashionMNIST-LeNet}} \\
    \midrule
    \textit{RA}($\uparrow$)    & \textbf{0.99 } & 0.99  & 0.99  & \textbf{0.99 } & 0.09  & 0.99 & \textbf{0.99}  & \textit{0A($\uparrow$)}    & 0.99  & \textbf{1.00 } \\
    \textit{FA}($\downarrow$)    & \textbf{0.00 } & 0.00  & 0.00  & \textbf{0.00 } & 0.11  & 0.00 & \textbf{0.00}  &\textit{ PS($\uparrow$)}    & 0.75 & \textbf{1.00}  \\
    \textit{ReA}($\downarrow$)   & \textbf{0.77 } & \underline{0.94}  & 0.96  & 0.97  & 0.96  & 1.00 & \textbf{1.00}  & \textit{ReA}  & \textbf{0.15 } & 0.70  \\
   \textit{ MIA}($\downarrow$)   & 0.85  & \textbf{0.47 } & 0.70  & \underline{0.68}  & 0.70  & \textbf{0.27 } & 0.99  & \textit{MIA}   & \textbf{0.53 } & 0.94  \\
   \textit{ Comp}($\downarrow$)  & 210.15  & 178.66  & 298.28  & \underline{158.96}  & \textbf{26.31 } & 213.60  & \textbf{81.94 } & \textit{Comp}  & 873.04  & \textbf{872.65 } \\
    \textit{Comm}($\downarrow$)  & 177K  & 177K  & 177K  & \textbf{11K} & 177K  & 177K  & \textbf{11K} & \textit{Comm}  & 177K  & \textbf{11K} \\
    \midrule
          & \multicolumn{10}{c}{\textit{Cifar10-ResNet18}} \\
    \midrule
    \textit{RA}($\uparrow$)    & \textbf{0.71 } & 0.67  & 0.65  & \underline{0.67}  & 0.64  & 0.73 & \textbf{0.73}  & \textit{0A($\uparrow$)}    & \textbf{0.56 } & 0.52  \\
    \textit{FA}($\downarrow$)    & \textbf{0.04 } & 0.04  & 0.05  & \underline{0.05}  & 0.06  & 0.00 & \textbf{0.00}  & \textit{PS($\uparrow$)}    & \textbf{0.55 } & 0.54  \\
    \textit{ReA}($\downarrow$)   & 0.49  & 0.48  & \textbf{0.41 } & \underline{0.42}  & 0.56  & 1.00 & \textbf{1.00}  & \textit{ReA}   & \textbf{1.00 } & 1.00  \\
    \textit{MIA}($\downarrow$)   & 0.78  & 0.67  & \textbf{0.43 } & \underline{0.65}  & 0.78  & \textbf{0.86 } & 0.96  & \textit{MIA}   & \textbf{0.98 } & 0.99  \\
   \textit{Comp}($\downarrow$)  & 434.39  & 990.91  & 1211.74  & \underline{262.20}  & \textbf{233.45 } & 735.07  & \textbf{183.02 } & \textit{Comp}  & 4619.82  & \textbf{1253.98 } \\
    \textit{Comm}($\downarrow$)  & 42.73M & 42.73M & 42.73M & \textbf{0.98M} & 42.73M & 42.73M & \textbf{0.98M} & \textit{Comm } & 42.73M & \textbf{0.98M} \\
    \midrule
          & \multicolumn{10}{c}{\textit{Cifar10-Transformer}} \\
    \midrule
    \textit{RA}($\uparrow$)    & 0.33  & 0.21  & 0.33  & \textbf{0.41 } & \underline{0.35}  & 0.27  & \textbf{0.44 } & \textit{0A($\uparrow$)}    & 0.05  & \textbf{0.33 } \\
    \textit{FA}($\downarrow$)    & \underline{0.08}  & 0.09  & 0.07  & \textbf{0.07 } & 0.07  & \textbf{0.00 } & 0.00  & \textit{PS($\uparrow$)}    & 0.20 & \textbf{0.54}  \\
    \textit{ReA}($\downarrow$)   & \underline{0.40 } & \textbf{0.25}  & 0.40  & 0.41  & 0.40  & 1.00  & \textbf{0.80 } &\textit{ReA}   & \textbf{0.03 } & 0.50  \\
    \textit{MIA}($\downarrow$)   & \underline{0.62}  & \textbf{0.41 } & 0.76  & \underline{0.62}  & 0.63  & \textbf{0.64 } & 0.88  & \textit{MIA}   & \textbf{0.85 } & 1.00  \\
    \textit{Comp}($\downarrow$)  & 5388.83  & \underline{1867.36}  & 4929.45  & \textbf{502.60 } & 3240.05  & 1207.56  & \textbf{218.30 } &\textit{ Comp}  & 628.44  & \textbf{342.14 } \\
    \textit{Comm}($\downarrow$)  & 36.21M & 36.21M & 36.21M & \textbf{0.71M} & 36.21M & 36.21M & \textbf{0.71M} &\textit{ Comm}  & 36.21M & \textbf{0.71M} \\
    \midrule
          & \multicolumn{10}{c}{\textit{Cifar100-ResNet18}} \\
    \midrule
   \textit{ RA}($\uparrow$)    & \textbf{0.39} & 0.19  & 0.25  & \underline{0.36}  & 0.35  & \textbf{0.34 } & 0.30  & \textit{0A($\uparrow$)}    & 0.57  & \textbf{0.79} \\
    \textit{FA}($\downarrow$)    & \underline{0.01}  & 0.01  & \textbf{0.00 } & \underline{0.01}  & 0.01  & 0.00 & \textbf{0.00}  & \textit{PS($\uparrow$)}    & 0.54 & \textbf{0.66}  \\
    \textit{ReA}($\downarrow$)   & 0.18  & \textbf{0.13 } & 0.17  & \underline{0.14}  & 0.20  & 1.00 & \textbf{1.00}  & \textit{ReA}   & 0.25  & \textbf{0.00 } \\
   \textit{ MIA}($\downarrow$)   & \underline{0.22}  & 0.28  & \textbf{0.08 } & 0.48  & 0.36  & 0.66 & \textbf{0.10} &\textit{ MIA}   & \textbf{0.98 } & 1.00  \\
    \textit{Comp}($\downarrow$)  & 443.86  & 1000.75  & 1598.55  &\underline{276.59}  & \textbf{235.65 } & 820.43  & \textbf{188.00 } & \textit{Comp}  & 4121.99  & \textbf{1580.91 } \\
    \textit{Comm}($\downarrow$)  & 42.91M & 42.91M & 42.91M & \textbf{0.98M} & 42.91M & 42.91M & \textbf{0.98M} & \textit{Comm } & 42.91M & \textbf{0.98M} \\
    \bottomrule
    \end{tabular}}
\caption{Main Results. ``0A" represents the accuracy of class 0, while ``PS" refers to the precision of the predicted class 0. The symbol $\uparrow$ indicates higher values are better, while $\downarrow$ indicates the opposite. ``E-F" is the short for Exact-Fun, and ``E-C" is EraseClient.}
  \label{results}%
\end{table*}%

\subsection{Critical layer identification}
\begin{figure}[!ht]
	\centering
	\begin{subfigure}[b]{0.48\linewidth}
		\includegraphics[width=\linewidth]{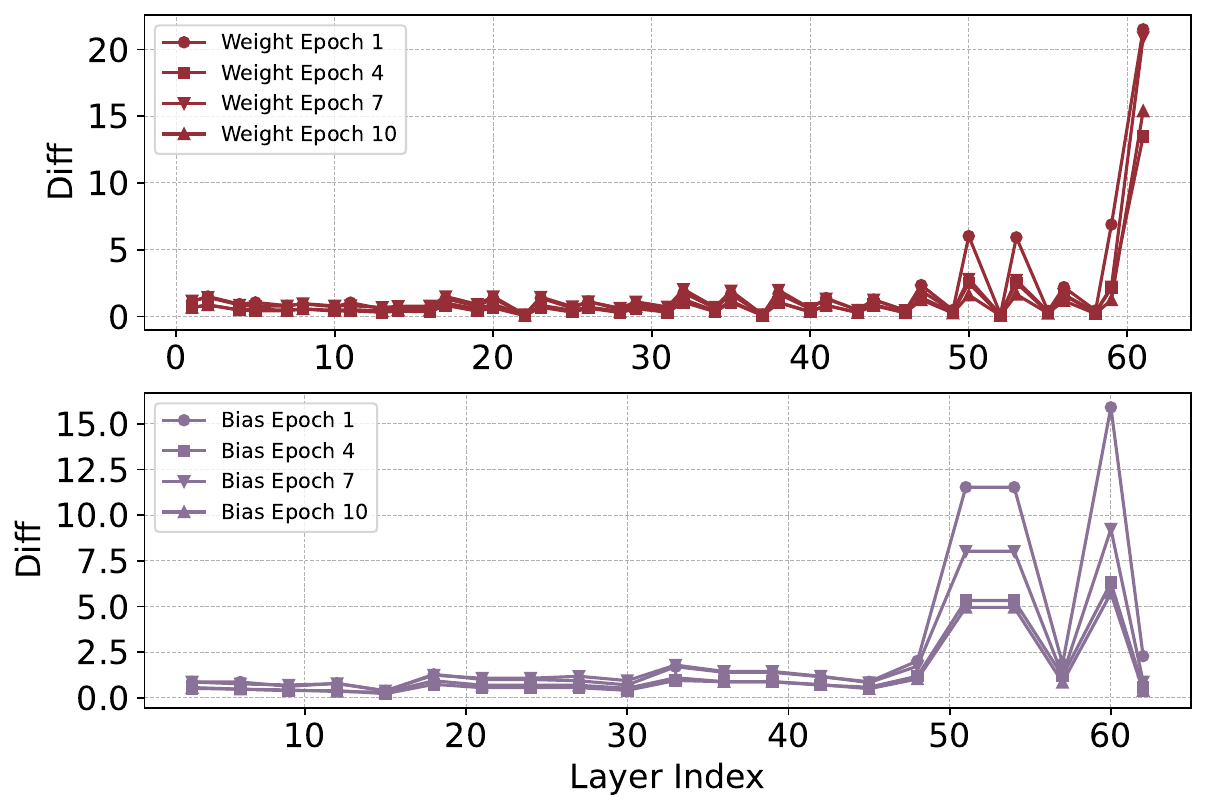}
		\caption{ResNet18}
		\label{param_change_cifar10_distri_0.1_model_resnet}
	\end{subfigure}
	\hfill
	\begin{subfigure}[b]{0.48\linewidth}
		\includegraphics[width=\linewidth]{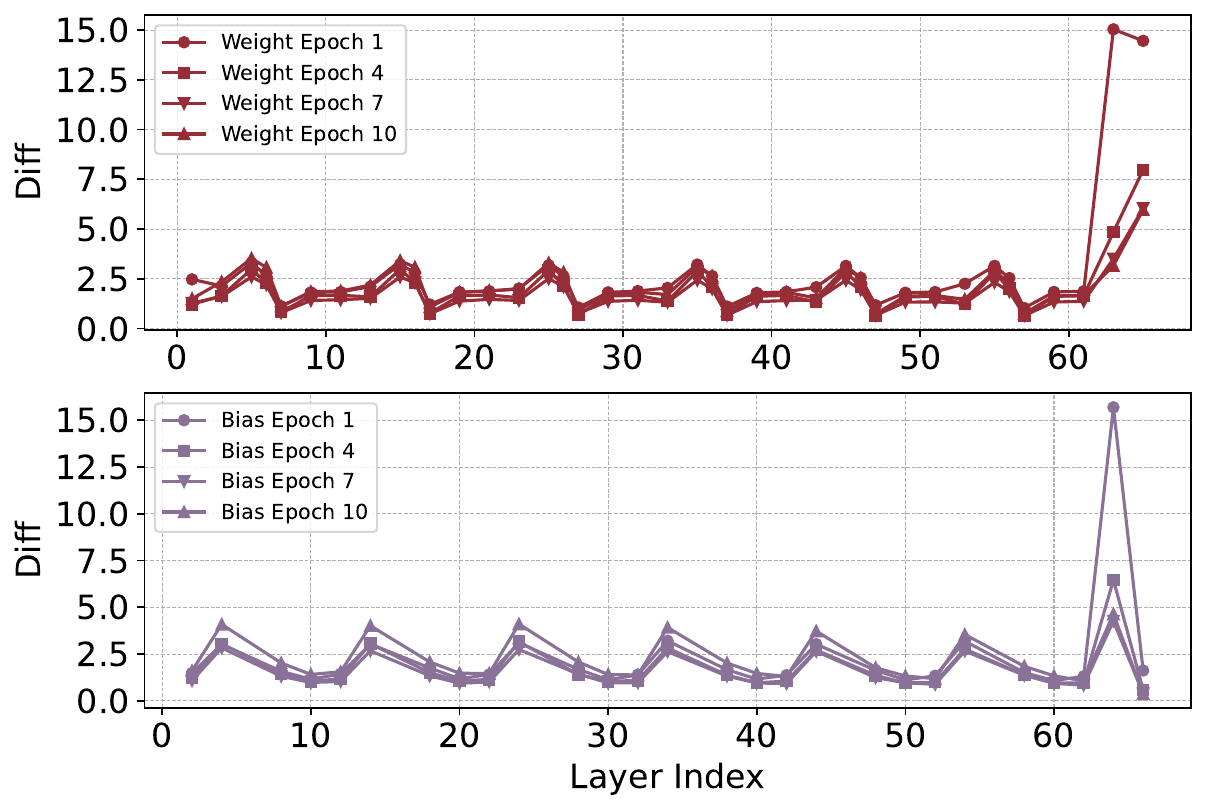}
		\caption{Transformer}
		\label{param_change_cifar100_distri_0.1_model_transformer}
	\end{subfigure}
	\caption{The average difference between local models and the server model across different models.}
	\label{param change}
\end{figure}

Before conducting unlearning, it is essential to identify the layers that are sensitive to knowledge. We segment the client data using a Dirichlet distribution with a parameter $\alpha$ of 0.1 to enhance knowledge disparity among clients. For the Cifar10 and Cifar100 datasets, we employ the ResNet18 and SimpleViT models, while the LeNet model is utilized for FashionMNIST. After obtaining locally trained models from different clients, we can observe the average change in each layer. In \cref{param change}, we present the Diff values for each layer of the ResNet18 and SimpleViT across different training iterations. We can see the last, second-to-last, sixth-to-last, and eighth-to-last layers of the ResNet18 model, and the last several layers of the Transformer model demonstrate heightened sensitivity to data variations across clients. Therefore, these layers will be designated as unlearning layers for sparse training in the subsequent unlearning process.

In fact, with the increasing number of federated iterations, the global model's knowledge generalization ability improves, leading to a gradual reduction in the gap of Diff values between layers. As illustrated in \cref{param change}, the gap is most pronounced when Epoch=1. However, as the number of iterations increases, this disparity decreases. Therefore, by comparing the model after a single federated iteration, it is possible to more precisely identify the critical layers sensitive to knowledge.

\subsection{Results Analysis}

\textbf{Unlearning performance.} In the client unlearning scenario, we use clients affected by Byzantine attacks as test cases. The mode of attack is label flipping, where one client's label is maliciously manipulated, resulting in a demand for unlearning. From \cref{results}, it can be observed that among the three metrics that directly measure forgetting effects—\textit{RA}, \textit{FA}, and \textit{ReA}—only \texttt{FUSED} is nearly on par with Rrtraining. This approach maintains a low accuracy on unlearned data while achieving a high accuracy on others, and even demonstrates overall superiority compared to Rrtraining, particularly in the Transformer model. It can be concluded that \texttt{FUSED} effectively unlearns the specified client knowledge while minimizing the impact on the original knowledge. This is attributable to the method proposed in this paper, which freezes the parameters of the original model and only trains the unlearning adapter, thereby avoiding direct modifications to the old knowledge and effectively reducing interference with the existing knowledge. Similarly, the same results are observed in class and sample unlearning; for further analysis, please refer to the Appendix.

\textbf{Knowledge interference.} To investigate the impact of unlearning on the overlapping knowledge across clients, we use the Cifar10 dataset, distributing 90\% of the data labeled as 0 and all data labeled as 1 to a client that needs to be forgotten. The remaining data, labeled from 2 to 9, and 10\% of the data labeled as 0, are randomly assigned to other clients. After unlearning, we evaluate the accuracy of the knowledge unique to the unlearning client (data labeled as 1), the accuracy of the overlapping knowledge (data labeled as 0 from the remaining clients), and the accuracy of the knowledge unique to the remaining clients (data labeled from 2 to 9). The final results are shown in \cref{KD-inter}. It can be observed that all methods completely forget the knowledge unique to the forgetting client, while only the \texttt{FUSED} method demonstrates improved performance on overlapping knowledge compared to Retraining. Therefore, \texttt{FUSED} can reduce knowledge interference.
\begin{table}[htbp]
  \centering
  \resizebox{\columnwidth}{!}{
    \begin{tabular}{cccccc}
    \toprule
    Method & Federaser & Retrain & E-F & FUSED & E-C \\
    \midrule
    F-Acc & 0.00  & 0.00  & 0.00  & 0.00  & 0.00  \\
    C-Acc & 0.14  & \textbf{0.38 } & 0.07  & \underline{0.37}  & 0.06  \\
    R-Acc & 0.27  & \underline{0.65}  & 0.64  & \underline{0.65}  & \textbf{0.66 } \\
    \bottomrule
    \end{tabular}}%
    \caption{``F-Acc" is the accuracy of the knowledge unique to the unlearning client, ``C-Acc" is for overlapping knowledge, and ``R-Acc" is for the knowledge unique to the remaining clients }
  \label{KD-inter}%
\end{table}%

\textbf{Unlearning cost.} In the unlearning process, resource overhead is an inevitable problem. \cref{results} primarily illustrates the consumption of computational and communication resources. Since the \texttt{FUSED} trains and transmits only the sparse adapters, it consistently demonstrates a significant advantage across nearly all unlearning scenarios and datasets. Additionally, in terms of storage resources, both Federaser and EraseClient require the retention of all client models and the global model during each round, which presents significant challenges regarding storage capacity. This demand increases exponentially with the number of clients and iterations, rendering it impractical in real-world applications. In contrast, \texttt{FUSED} only requires the storage of a complete global model and its adapters. Moreover, when compared to the retraining method, the retraining method achieves \textit{RA/FA} values of 0.71/0.04 when data is complete, and \texttt{FUSED} achieves \textit{RA/FA} values of 0.67/0.05. When we reduce the number of retraining data by half, \texttt{FUSED} maintains \textit{RA/FA} values of 0.65/0.03, indicating no significant decline in unlearning performance. This suggests that \texttt{FUSED} can achieve results comparable to retraining with less data, thereby conserving storage resources.

\textbf{Privacy protection.} When unlearned data is users' privacy, even if the model shows great unlearning performance, an attacker may still be able to discern which data corresponds to unlearned private information and which does not, particularly in the context of member inference attacks. Therefore, it is crucial to evaluate the privacy leakage rate of the model after unlearning. The \textit{MIA} values for \texttt{FUSED} are generally comparable to those of the Retraining method, and in most instances, they remain at a relatively low level. This indicates that \texttt{FUSED}'s capability to mitigate privacy leakage is on par with that of other methods.

\begin{figure}[!ht]
	\centering
	\includegraphics[width=\linewidth]{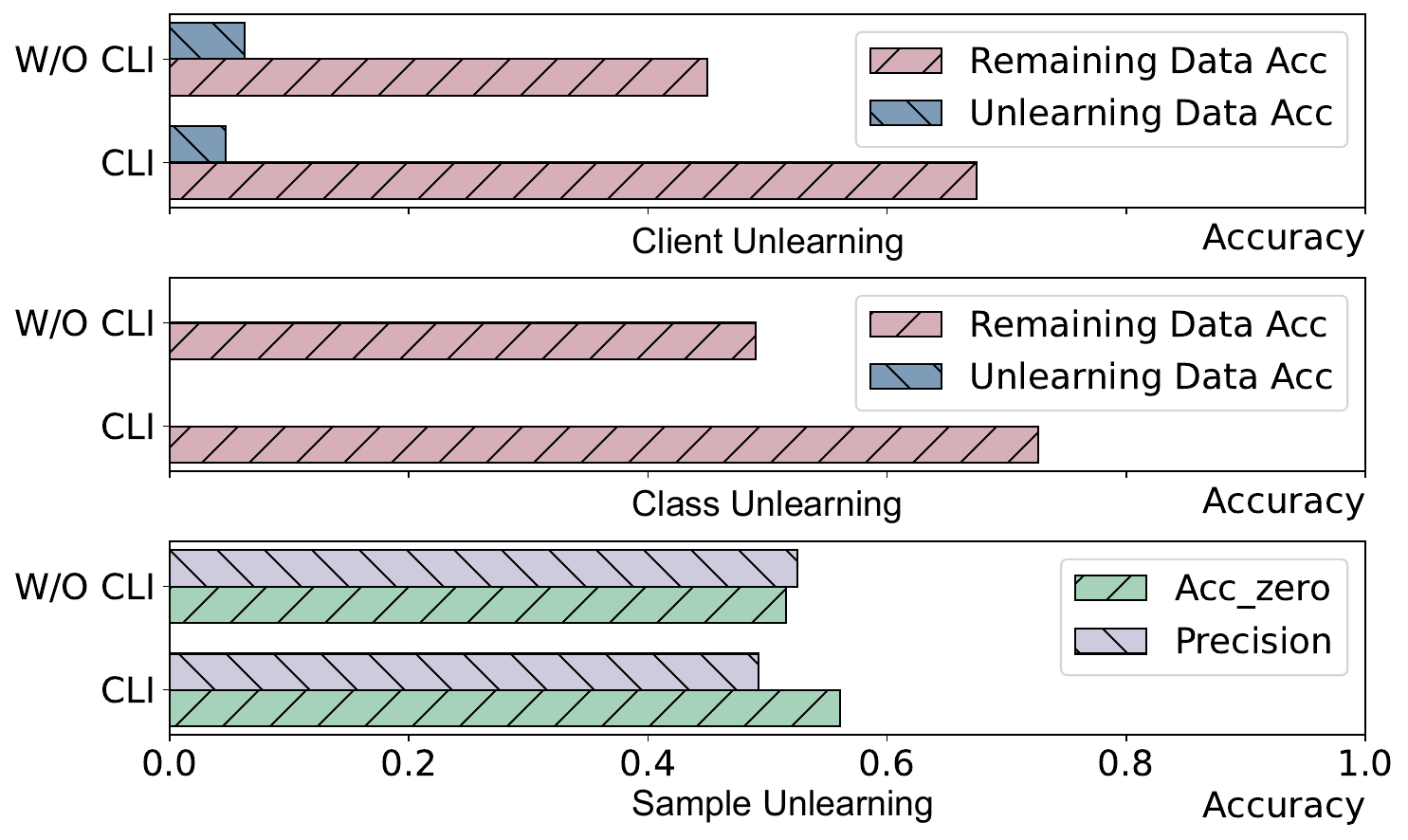}
	\caption{Ablation study of CLI.}
	\label{Ablation studies}
\end{figure}

\textbf{Ablation study.} To illustrate the necessity of CLI, we conduct an ablation study using the Cifar10 dataset, with the experimental results presented in \cref{Ablation studies}. In \cref{Ablation studies}, ``W/O CLI" denotes the effect of \texttt{FUSED} achieved by randomly selected layers. It is evident that, with the implementation of CLI, the accuracy of remaining knowledge is higher in both client unlearning and class unlearning scenarios. Although the disparity is smaller in sample unlearning, it still maintains a comparable level. This indicates that CLI can more accurately identify the model layers that are more sensitive to knowledge, thereby enhancing the unlearning effect.

%% file: sec/6_conclusion.tex
\section{Conclusion and Discussion}
\textbf{Conclusion.} This paper focuses on the problem of unlearning within FL. To address the challenges of indiscriminate unlearning, irreversible unlearning, and significant unlearning costs, we propose a reversible federated unlearning method via selective sparse adapters (\texttt{FUSED}). Firstly, by comparing the client model with the server model, we identify critical layers to unlearn. Then, independent sparse unlearning adapters are constructed for each unlearning layer. After that, only the sparse adapters are retrained, achieving efficient resource utilization. In this way, \texttt{FUSED} greatly reduces the knowledge interference. Furthermore, independent adapters are easy to remove to facilitate memory recovery. Finally, we validate \texttt{FUSED} in client unlearning scenarios based on Byzantine attacks, sample unlearning scenarios based on backdoor attacks, and class unlearning scenarios. The results show that \texttt{FUSED}'s unlearning effectiveness matches that of Retraining, surpassing other baselines while significantly reducing costs.

\textbf{Discussion.} In addition to unlearning, the proposed adapters can also serve as knowledge editors, adjusting the model's knowledge on different occasions. For instance, they can help unlearn private information and overcome catastrophic forgetting simultaneously. Moreover, when the knowledge editing requirements vary among clients, combinations of adapters can enhance global generalization. However, there are some limitations of \texttt{FUSED} we can not overlook, for example, it still requires a great number of remaining data to train the adapters. Some techniques like data compression are expected to solve this problem. Meanwhile, compared to some methods that only adjust parameters on the server, the FUSED method, which is based on retraining, requires the participation of all clients that contain residual knowledge, demanding a higher level of client engagement.

%% file: main.bbl
\begin{thebibliography}{46}
\providecommand{\natexlab}[1]{#1}
\providecommand{\url}[1]{\texttt{#1}}
\expandafter\ifx\csname urlstyle\endcsname\relax
  \providecommand{\doi}[1]{doi: #1}\else
  \providecommand{\doi}{doi: \begingroup \urlstyle{rm}\Url}\fi

\bibitem[Baumhauer et~al.(2022)Baumhauer, Sch{\"o}ttle, and Zeppelzauer]{Baumhauer2022}
Thomas Baumhauer, Pascal Sch{\"o}ttle, and Matthias Zeppelzauer.
\newblock Machine unlearning: Linear filtration for logit-based classifiers.
\newblock \emph{Machine Learning}, 111\penalty0 (9):\penalty0 3203--3226, 2022.

\bibitem[Bourtoule et~al.(2021)Bourtoule, Chandrasekaran, Choquette-Choo, Jia, Travers, Zhang, Lie, and Papernot]{Bourtoule2021}
Lucas Bourtoule, Varun Chandrasekaran, Christopher~A Choquette-Choo, Hengrui Jia, Adelin Travers, Baiwu Zhang, David Lie, and Nicolas Papernot.
\newblock Machine unlearning.
\newblock In \emph{2021 IEEE Symposium on Security and Privacy}, pages 141--159. IEEE, 2021.

\bibitem[Brophy and Lowd(2021)]{Brophy2021}
Jonathan Brophy and Daniel Lowd.
\newblock Machine unlearning for random forests.
\newblock In \emph{International Conference on Machine Learning}, pages 1092--1104. PMLR, 2021.

\bibitem[Cao et~al.(2023)Cao, Jia, Zhang, and Gong]{Cao2023}
Xiaoyu Cao, Jinyuan Jia, Zaixi Zhang, and Neil~Zhenqiang Gong.
\newblock Fedrecover: Recovering from poisoning attacks in federated learning using historical information.
\newblock In \emph{2023 IEEE Symposium on Security and Privacy}, pages 1366--1383. IEEE, 2023.

\bibitem[Che et~al.(2023)Che, Zhou, Zhang, Lyu, Liu, Yan, Dou, and Huan]{Che2023}
Tianshi Che, Yang Zhou, Zijie Zhang, Lingjuan Lyu, Ji Liu, Da Yan, Dejing Dou, and Jun Huan.
\newblock Fast federated machine unlearning with nonlinear functional theory.
\newblock In \emph{International conference on machine learning}, pages 4241--4268. PMLR, 2023.

\bibitem[Chen et~al.(2022{\natexlab{a}})Chen, Sun, Zhang, and Ding]{Chen2022}
Chong Chen, Fei Sun, Min Zhang, and Bolin Ding.
\newblock Recommendation unlearning.
\newblock In \emph{Proceedings of the ACM Web Conference 2022}, pages 2768--2777, 2022{\natexlab{a}}.

\bibitem[Chen et~al.(2024)Chen, Wang, Mi, Liu, Wang, Ren, and Shen]{Chen2024}
Kongyang Chen, Zixin Wang, Bing Mi, Waixi Liu, Shaowei Wang, Xiaojun Ren, and Jiaxing Shen.
\newblock Machine unlearning in large language models, 2024.

\bibitem[Chen et~al.(2022{\natexlab{b}})Chen, Zhang, Wang, Backes, Humbert, and Zhang]{Chen2022a}
Min Chen, Zhikun Zhang, Tianhao Wang, Michael Backes, Mathias Humbert, and Yang Zhang.
\newblock Graph unlearning.
\newblock In \emph{Proceedings of the 2022 ACM SIGSAC Conference on Computer and Communications Security}, pages 499--513, 2022{\natexlab{b}}.

\bibitem[Chowdhury et~al.(2024)Chowdhury, Choromanski, Sehanobish, Dubey, and Chaturvedi]{Chowdhury2024}
Somnath Basu~Roy Chowdhury, Krzysztof Choromanski, Arijit Sehanobish, Avinava Dubey, and Snigdha Chaturvedi.
\newblock Towards scalable exact machine unlearning using parameter-efficient fine-tuning.
\newblock \emph{arXiv preprint arXiv:2406.16257}, 2024.

\bibitem[{CNN}(2023)]{CNN2023}
{CNN}.
\newblock New york times sues openai and microsoft, 2023.
\newblock [Online; accessed: 2024-08-02].

\bibitem[Felps et~al.(2020)Felps, Schwickerath, Williams, Vuong, Briggs, Hunt, Sakmar, Saranchak, and Shumaker]{Felps2020}
Daniel~L. Felps, Amelia~D. Schwickerath, Joyce~D. Williams, Trung~N. Vuong, Alan Briggs, Matthew Hunt, Evan Sakmar, David~D. Saranchak, and Tyler Shumaker.
\newblock Class clown: Data redaction in machine unlearning at enterprise scale, 2020.

\bibitem[Garg et~al.(2020)Garg, Goldwasser, and Vasudevan]{Garg2020}
Sanjam Garg, Shafi Goldwasser, and Prashant~Nalini Vasudevan.
\newblock Formalizing data deletion in the context of the right to be forgotten.
\newblock In \emph{Annual International Conference on the Theory and Applications of Cryptographic Techniques}, pages 373--402. Springer, 2020.

\bibitem[Golatkar et~al.(2020)Golatkar, Achille, and Soatto]{Golatkar2020}
Aditya Golatkar, Alessandro Achille, and Stefano Soatto.
\newblock Eternal sunshine of the spotless net: Selective forgetting in deep networks.
\newblock In \emph{Proceedings of the IEEE/CVF Conference on Computer Vision and Pattern Recognition}, pages 9304--9312, 2020.

\bibitem[Graves et~al.(2021)Graves, Nagisetty, and Ganesh]{Graves2021}
Laura Graves, Vineel Nagisetty, and Vijay Ganesh.
\newblock Amnesiac machine learning.
\newblock In \emph{Proceedings of the AAAI Conference on Artificial Intelligence}, pages 11516--11524, 2021.

\bibitem[Gu et~al.(2024)Gu, Zhu, Zhang, Zhao, Han, Fan, and Yang]{Gu2024}
Hanlin Gu, Gongxi Zhu, Jie Zhang, Xinyuan Zhao, Yuxing Han, Lixin Fan, and Qiang Yang.
\newblock Unlearning during learning: An efficient federated machine unlearning method, 2024.

\bibitem[Guo et~al.(2023)Guo, Goldstein, Hannun, and van~der Maaten]{Guo2019}
Chuan Guo, Tom Goldstein, Awni Hannun, and Laurens van~der Maaten.
\newblock Certified data removal from machine learning models, 2023.

\bibitem[Guo et~al.(2020)Guo, Wang, Li, and Yan]{Guo2020}
Shaopeng Guo, Yujie Wang, Quanquan Li, and Junjie Yan.
\newblock Dmcp: Differentiable markov channel pruning for neural networks.
\newblock In \emph{Proceedings of the IEEE/CVF Conference on Computer Vision and Pattern Recognition}, pages 1539--1547, 2020.

\bibitem[Halimi et~al.(2022)Halimi, Kadhe, Rawat, and Baracaldo]{Halimi2022}
Anisa Halimi, Swanand Kadhe, Ambrish Rawat, and Nathalie Baracaldo.
\newblock Federated unlearning: How to efficiently erase a client in fl?
\newblock In \emph{International Conference on Machine Learning}. PMLR, 2022.

\bibitem[He et~al.(2021)He, Wang, Cao, Ding, Chen, Feng, Wang, and Huang]{He2021}
Dongxiao He, Youyou Wang, Jinxin Cao, Weiping Ding, Shizhan Chen, Zhiyong Feng, Bo Wang, and Yuxiao Huang.
\newblock A network embedding-enhanced bayesian model for generalized community detection in complex networks.
\newblock \emph{Information Sciences}, 575:\penalty0 306--322, 2021.

\bibitem[Izzo et~al.(2021)Izzo, Smart, Chaudhuri, and Zou]{Izzo2021}
Zachary Izzo, Mary~Anne Smart, Kamalika Chaudhuri, and James Zou.
\newblock Approximate data deletion from machine learning models.
\newblock In \emph{International Conference on Artificial Intelligence and Statistics}, pages 2008--2016. PMLR, 2021.

\bibitem[Krizhevsky(2009)]{Krizhevsky2009}
Alex Krizhevsky.
\newblock Cifar-10 and cifar-100 datasets.
\newblock http://www.cs.toronto.edu/~kriz/cifar.html, 2009.
\newblock Accessed: 2024-08-14.

\bibitem[Lee et~al.(2019)Lee, Xiao, Schoenholz, Bahri, Novak, Sohl{-}Dickstein, and Pennington]{Lee2019}
Jaehoon Lee, Lechao Xiao, Samuel~S. Schoenholz, Yasaman Bahri, Roman Novak, Jascha Sohl{-}Dickstein, and Jeffrey Pennington.
\newblock Wide neural networks of any depth evolve as linear models under gradient descent.
\newblock In \emph{NeurIPS}, pages 8570--8581, 2019.

\bibitem[Li et~al.(2024)Li, Hsu, Chen, and Marculescu]{Li2024}
Guihong Li, Hsiang Hsu, Chun-Fu Chen, and Radu Marculescu.
\newblock Fast-ntk: Parameter-efficient unlearning for large-scale models.
\newblock In \emph{Proceedings of the IEEE/CVF Conference on Computer Vision and Pattern Recognition}, pages 227--234, 2024.

\bibitem[Liu et~al.(2020)Liu, Ma, Yang, Wang, and Liu]{Liu2020}
Gaoyang Liu, Xiaoqiang Ma, Yang Yang, Chen Wang, and Jiangchuan Liu.
\newblock Federated unlearning.
\newblock \emph{arXiv preprint arXiv:2012.13891}, 2020.

\bibitem[Liu et~al.(2021)Liu, Ma, Yang, Wang, and Liu]{Liu2021}
Gaoyang Liu, Xiaoqiang Ma, Yang Yang, Chen Wang, and Jiangchuan Liu.
\newblock Federaser: Enabling efficient client-level data removal from federated learning models.
\newblock In \emph{2021 IEEE/ACM 29th International Symposium on Quality of Service}, pages 1--10. IEEE, 2021.

\bibitem[Liu et~al.(2024)Liu, Yao, Jia, Casper, Baracaldo, Hase, Yao, Liu, Xu, Li, Varshney, Bansal, Koyejo, and Liu]{Liu2024}
Sijia Liu, Yuanshun Yao, Jinghan Jia, Stephen Casper, Nathalie Baracaldo, Peter Hase, Yuguang Yao, Chris~Yuhao Liu, Xiaojun Xu, Hang Li, Kush~R. Varshney, Mohit Bansal, Sanmi Koyejo, and Yang Liu.
\newblock Rethinking machine unlearning for large language models, 2024.

\bibitem[Liu et~al.(2022)Liu, Xu, Yuan, Wang, and Li]{Liu2022}
Yi Liu, Lei Xu, Xingliang Yuan, Cong Wang, and Bo Li.
\newblock The right to be forgotten in federated learning: An efficient realization with rapid retraining.
\newblock In \emph{IEEE INFOCOM 2022-IEEE Conference on Computer Communications}, pages 1749--1758. IEEE, 2022.

\bibitem[McMahan et~al.(2017)McMahan, Moore, Ramage, Hampson, and y~Arcas]{McMahan2017}
Brendan McMahan, Eider Moore, Daniel Ramage, Seth Hampson, and Blaise~Aguera y Arcas.
\newblock Communication-efficient learning of deep networks from decentralized data.
\newblock In \emph{Artificial Intelligence and Statistics}, pages 1273--1282. PMLR, 2017.

\bibitem[Nguyen et~al.(2022)Nguyen, Huynh, Nguyen, Liew, Yin, and Nguyen]{Nguyen2022}
Thanh~Tam Nguyen, Thanh~Trung Huynh, Phi~Le Nguyen, Alan Wee-Chung Liew, Hongzhi Yin, and Quoc Viet~Hung Nguyen.
\newblock A survey of machine unlearning, 2022.

\bibitem[Ren et~al.(2021)Ren, Sun, and Wei]{Ren2021}
Zhenwen Ren, Quansen Sun, and Dong Wei.
\newblock Multiple kernel clustering with kernel k-means coupled graph tensor learning.
\newblock In \emph{Proceedings Of the AAAI Conference on Artificial Intelligence}, pages 9411--9418, 2021.

\bibitem[Schelter et~al.(2021)Schelter, Grafberger, and Dunning]{Schelter2021}
Sebastian Schelter, Stefan Grafberger, and Ted Dunning.
\newblock Hedgecut: Maintaining randomised trees for low-latency machine unlearning.
\newblock In \emph{Proceedings of the 2021 International Conference on Management of Data}, pages 1545--1557, 2021.

\bibitem[Sinha et~al.(2020)Sinha, Pal, and De]{Sinha2020}
Rituparna Sinha, Rajat~K Pal, and Rajat~K De.
\newblock Genseg and mr-genseg: A novel segmentation algorithm and its parallel mapreduce based approach for identifying genomic regions with copy number variations.
\newblock \emph{IEEE/ACM Transactions on Computational Biology and Bioinformatics}, 19\penalty0 (1):\penalty0 443--454, 2020.

\bibitem[Su and Li(2023)]{Su2023}
Ningxin Su and Baochun Li.
\newblock Asynchronous federated unlearning.
\newblock In \emph{IEEE INFOCOM 2023-IEEE Conference on Computer Communications}, pages 1--10. IEEE, 2023.

\bibitem[Tarun et~al.(2023)Tarun, Chundawat, Mandal, and Kankanhalli]{Tarun2023}
Ayush~K Tarun, Vikram~S Chundawat, Murari Mandal, and Mohan Kankanhalli.
\newblock Fast yet effective machine unlearning.
\newblock \emph{IEEE Transactions on Neural Networks and Learning Systems}, 2023.

\bibitem[Villaronga et~al.(2018)Villaronga, Kieseberg, and Li]{Villaronga2018}
Eduard~Fosch Villaronga, Peter Kieseberg, and Tiffany Li.
\newblock Humans forget, machines remember: Artificial intelligence and the right to be forgotten.
\newblock \emph{Computer Law \& Security Review}, 34\penalty0 (2):\penalty0 304--313, 2018.

\bibitem[Wang et~al.(2022)Wang, Guo, Xie, and Qi]{Wang2022}
Junxiao Wang, Song Guo, Xin Xie, and Heng Qi.
\newblock Federated unlearning via class-discriminative pruning.
\newblock In \emph{Proceedings of the ACM Web Conference 2022}, pages 622--632, 2022.

\bibitem[Wang et~al.(2023)Wang, Tian, Zhang, Liu, and Yu]{Wang2023}
Weiqi Wang, Zhiyi Tian, Chenhan Zhang, An Liu, and Shui Yu.
\newblock Bfu: Bayesian federated unlearning with parameter self-sharing.
\newblock In \emph{Proceedings of the 2023 ACM Asia Conference on Computer and Communications Security}, pages 567--578, 2023.

\bibitem[Warnecke et~al.(2023)Warnecke, Pirch, Wressnegger, and Rieck]{Warnecke2021}
Alexander Warnecke, Lukas Pirch, Christian Wressnegger, and Konrad Rieck.
\newblock Machine unlearning of features and labels, 2023.

\bibitem[Wu et~al.(2022{\natexlab{a}})Wu, Zhu, and Mitra]{Wu2022a}
Chen Wu, Sencun Zhu, and Prasenjit Mitra.
\newblock Federated unlearning with knowledge distillation, 2022{\natexlab{a}}.

\bibitem[Wu et~al.(2022{\natexlab{b}})Wu, Guo, Wang, Hong, Zhang, and Ding]{Wu2022}
Leijie Wu, Song Guo, Junxiao Wang, Zicong Hong, Jie Zhang, and Yaohong Ding.
\newblock Federated unlearning: Guarantee the right of clients to forget.
\newblock \emph{IEEE Network}, 36\penalty0 (5):\penalty0 129--135, 2022{\natexlab{b}}.

\bibitem[Wu et~al.(2020)Wu, Dobriban, and Davidson]{Wu2020}
Yinjun Wu, Edgar Dobriban, and Susan Davidson.
\newblock Deltagrad: Rapid retraining of machine learning models.
\newblock In \emph{International Conference on Machine Learning}, pages 10355--10366. PMLR, 2020.

\bibitem[Xiao et~al.(2017)Xiao, Rasul, and Vollgraf]{Xiao2017}
Han Xiao, Kashif Rasul, and Roland Vollgraf.
\newblock Fashion-mnist: a novel image dataset for benchmarking machine learning algorithms, 2017.

\bibitem[Xiong et~al.(2023)Xiong, Li, Li, and Cai]{Xiong2023}
Zuobin Xiong, Wei Li, Yingshu Li, and Zhipeng Cai.
\newblock Exact-fun: An exact and efficient federated unlearning approach.
\newblock In \emph{2023 IEEE International Conference on Data Mining}, pages 1439--1444. IEEE, 2023.

\bibitem[Yoshioka(2024)]{Yoshioka2024}
Kentaro Yoshioka.
\newblock vision-transformers-cifar10: Training vision transformers (vit) and related models on cifar-10.
\newblock \url{https://github.com/kentaroy47/vision-transformers-cifar10}, 2024.

\bibitem[Zhang et~al.(2023)Zhang, Nakamura, Isohara, and Sakurai]{Zhang2023}
Haibo Zhang, Toru Nakamura, Takamasa Isohara, and Kouichi Sakurai.
\newblock A review on machine unlearning.
\newblock \emph{SN Computer Science}, 4\penalty0 (4):\penalty0 337, 2023.

\bibitem[Zhang et~al.(2022)Zhang, Bai, Huang, and Xu]{Zhang2022}
Peng-Fei Zhang, Guangdong Bai, Zi Huang, and Xin-Shun Xu.
\newblock Machine unlearning for image retrieval: A generative scrubbing approach.
\newblock In \emph{Proceedings of the 30th ACM International Conference on Multimedia}, pages 237--245, 2022.

\end{thebibliography}
